\documentclass[a4paper, 10pt]{article}
\usepackage[english]{babel}  

\usepackage{cite}

\usepackage{cite}
\usepackage{amsmath,amssymb,amsfonts}
 \usepackage{algorithmic}
\usepackage{graphicx}
\usepackage{textcomp}
\usepackage{xcolor}
\def\BibTeX{{\rm B\kern-.05em{\sc i\kern-.025em b}\kern-.08em
    T\kern-.1667em\lower.7ex\hbox{E}\kern-.125emX}}

\usepackage[utf8]{inputenc}
\usepackage{times}
\usepackage{soul}
\usepackage{url}
\usepackage[hidelinks]{hyperref}
\usepackage[utf8]{inputenc}
\usepackage[small]{caption}
\usepackage{booktabs}
\usepackage{algorithm}

\usepackage{nomencl}
\makenomenclature

\usepackage{hyperref,color}    
\usepackage[hypcap]{caption}  

\newcommand{\reportTitle}{Bootstrapping confidence in future safety\\based on past safe operation}

\newcommand{\manualDate}{20 October 2021}

\usepackage{fancyhdr}
\begin{document}
%


\title{ \reportTitle }
\date{\manualDate}


\author{Peter Bishop$^{1,2}$, Andrey Povyakalo$^{1}$ and Lorenzo Strigini$^{1}$ 
\thanks{This work was supported in part by ICRI-SAVe, the Intel Collaborative Research Institute on Safe Automated Vehicles.}
}

\date{ $^1$Centre for Software Reliability\\  
 City, University of London, United Kingdom
\\
 {\tt\small \{P.Bishop,A.A.Povyakalo,L.Strigini\}@city.ac.uk}
   \\ $^2$Adelard LLP, London, United Kingdom\\
 {\tt\small pgb@adelard.com}  \\[2ex]%
\manualDate}


%


\maketitle

\begin{abstract}
With autonomous vehicles (AVs), a major concern is the inability to give meaningful quantitative assurance of safety, to the extent required by society -- e.g. that an AV must be at least as safe as a good human driver -- before that AV is in extensive use.
We demonstrate an approach to achieving more moderate, but useful, confidence, e.g., confidence of low enough probability of causing accidents in the early phases of operation.
This formalises mathematically the common approach of operating a system on a  limited basis in the hope that mishap-free operation will confirm one's confidence in its safety and allow progressively more extensive operation:
a process of ``bootstrapping'' of confidence.
Translating that intuitive approach into theorems shows: (1) that it is substantially sound in the right circumstances, and could be a good method for deciding about the early deployment phase for an AV; (2) how much confidence can be rightly derived from such a ``cautious deployment'' approach, so that we can avoid over-optimism; (3) under which conditions our sound formulas for future confidence are applicable; (4) thus, which analyses of the concrete situations, and/or constraints on practice, are needed in order to enjoy the advantages of provably correct confidence in adequate future safety.
\end{abstract}


\section{Introduction}

A major concern for developers of autonomous vehicle, authorities responsible for authorising their use, and end users, is how to achieve confidence that they will be safe enough in operation.
The safety objectives are inevitably, at least in part, quantitative: we do not want accidents to be too frequent.
This might be stated as a specific maximum frequency of accidents or fatalities per mile, or a requirement that these frequencies be no greater than some fraction of the average for human drivers, or comparable to the better human drivers, etc..
It is important to have quantitative arguments that properly demonstrate what level of confidence we can have in these requirements being satisfied.

However, a major difficulty arises when spelling out these quantitative requirements for road vehicles on public roads:
when these reasonable requirements are translated into numerical targets about, for instance, a low enough ``probability of fatal accidents per mile driven'', demonstrating that the requirement is satisfied by just operating the vehicles and collecting statistical evidence is impractical.
The cost and practical difficulties in building such confidence statistically seem unsurmountable \cite{kalra_driving_2016}, as demonstrated previously by other authors \cite{LittlewoodStriginiCACM93, ButlerFinelliInfeasibility1993}.
Even when one takes into account evidence accumulated prior to operation or testing -- evidence of precautions taken in development and in the system design, as is normal for safety-critical systems -- one would need this evidence to support such strong confidence that few would consider the argument believable  \cite{zhao_assessing_2020}, especially for functions based on machine learning.

The problem is not that the mathematical methods are faulty, but that these very low rates of accidents, although reasonable to require, are hard to believe without large amounts of empirical evidence.

It turns out that an alternative useful quantitative claim can be more easily supported, about the probability of safety over a finite amount of future operation \cite{strigini_software_2013}.

That is, the claim supported is not that the probability per mile is less than a certain number, but that the probability of a mishap occurring during a period of future operation is less than a certain probability. 
For instance, for a nuclear reactor protection system or a critical flight control function, the practical requirement is that causing accidents must be unlikely over the whole lifetime of the type. For a vehicle which is being operated so as to gain confidence to warrant further operation, this requirement might be that accidents be unlikely over the next few months or years of operation.
The claim could be that for a fleet of cars there is a 90\% probability that it will not suffer accidents due to its self-driving functions, over the next year of operation.
In this paper, we demonstrate how this approach can support confidence in safety in some scenarios, including the early adoption stage for a new vehicle.

Many different kinds of AVs are in development or in use, creating many different scenarios for this problem of gaining confidence in their safety. We will refer to two types of AVs that are at opposite extremes of the range, from the viewpoint of how easy or hard it is to accumulate confidence in their future safety from progressively extended periods of operation.
At one end of the spectrum, we will call ``A-type'' vehicles those ``SAE level 5'' AVs meant to be sold to many millions of ordinary consumers and to transport them over public roads, in complex and quite unpredictable environments. At the opposite end, we will call ``Z-type'' those AVs that are deployed in small numbers to perform well-understood, limited tasks in constrained and protected environments, e.g., a self-moving crane in a factory or a heavy truck in an open-face mine.

Z-type AVs may still have extreme safety requirements (e.g. if they transport dangerous material in a chemical plant),
but the simpler environment and less pressure for high performance and fast evolution reduce difficulties in both their development and assessment.

\section{Basic model and results}

We show the reasoning for the case that the process of mishaps occurring\footnote{We use the word \emph{mishap} as an umbrella term for the negative event about which one wants to give predictions. Some may want to reason about probability of a deadly accident as in \cite{kalra_driving_2016, zhao_assessing_2020}, or of any accident, or of any \emph{potential} accident, e.g. any violation of an assigned safety envelope, or even a  failure that analyses reveal \emph{could} cause an accident.
In safety, this last definition (\emph{potential} accident) is typically the one would want to use; but to have data for statistical inference, it may be necessary to reason just about serious accidents, those that are reported and will appear in logs.}
can be modelled as ``Bernoulli trials'': the system is subjected to a series of \emph{demands}, and mishaps on different demands are independent events with the same probability (\emph{probability of failure per demand},  \emph{pfd}).
Bernoulli processes are a common model for failure processes \cite{iec_61508_2010, NUREG6823atwood2003handbook}.
One could call ``demand'' a single trip; or, as in \cite{kalra_driving_2016,zhao_assessing_2020}, driving a mile or a kilometre. If mishaps are rare, the Bernoulli process should be a tolerable approximation of reality (see \cite{zhao_assessing_2020} for a discussion and references), despite the fact that successive demands, if defined this way, are not independent \cite{strigini_testing_1996}.

Our proposed way of reasoning used Bayesian inference: the uncertain values of interest are considered random variables, with a ``prior'' distribution that represents the state of knowledge and uncertainty about their values before new evidence is observed, which is updated  on the basis of this new evidence.
In this context, the new evidence is that some amount of operation was completed with no mishaps.
In particular, we used the approach that we call ``conservative Bayesian inference'': we avoid the need to specify the prior distribution in full, and instead depend on specifying only some characteristics of it that one \emph{can} trust to have reasons for believing.
This approach is detailed elsewhere (e.g. \cite[section 3]{Zhao2020_AVfromISSRE_NWTES}, for a general description; \cite{strigini_software_2013} for the specific models applied in the present paper).
These details do not matter here; the essence of our previous results is that, on the basis of a very limited partial description of the prior distribution, they allowed one to state rigorously, after observing a period of mishap-free operation, what confidence could then have ``as a  minimum'' that the system would not suffer mishaps over a certain future amount of operation.

Our premises were: 

\begin{itemize}
\item the real question of interest is whether the risk of having \emph{any} mishap \emph{over a certain period of operation} (e.g., the whole lifetime of the system; or the next year of operation) is acceptably low (in other cases, whether the probability of \emph{too many} mishaps is); estimating a \emph{pfd} is just a mathematical detail of how one can answer this question;

\item for many systems, there is strong confidence, \emph{before} we start operating them, that they are sufficiently safe.
Indeed, those deciding to start operation of such systems would not do so if they did not have this confidence.
The source of the confidence is typically in that these systems were developed following good quality practice, the code was extensively verified, etc.
How much confidence this evidence \emph{should} really generate is a separate problem.
\end{itemize}

This prior confidence is affected by uncertainty, of course. Expensively developed and verified systems have been put into operation despite failure modes with astonishing high probabilities: historically, e.g., the early Space Shuttle software had a probability of 1 in 67 of per flight of failing to start; the initial version of the Ariane 5's control system, a 100\% probability of destroying the rocket.
In general, a thorough interrogation of what one knows should indicate some estimate of a probability (less than 100\%) that the \emph{pfd} is acceptably low.
We note that this is a Bayesian probability: the system's \emph{pfd}, given the way the system will be used and the world around it, is a specific number, but is unknown.
This probability of acceptable \emph{pfd} describes our ``epistemic'' uncertainty about what the \emph{pfd}'s real value is; this uncertainty is a crucial factor in our decision whether to take the ``gamble'' of operating the system.

In Bayesian terms, the unknown \emph{pfd} is a random variable -- we will call it $Q$ for brevity -- and our uncertainty about it is described by a  probability distribution, say a probability density function, $f_{{Q}} \left( q \right) $.

\subsection{Argument based on probability of \emph{pfd}$=0$}
The simplest version of our conservative form of reasoning applies for systems so simple that one has some substantial confidence that they are \emph{free} from safety-relevant faults, i.e., that their \emph{pfd} is zero.\footnote{This case is convenient for the purpose of presentation since its results yield simpler plots than the general case, which we introduce in section \ref{subsec_Effective_fault-freeness}, of imperfect confidence in the  \emph{pfd} being lower than a small, non-zero bound.}

Of course one never has 100\% confidence of this. Thus a parameter of this kind of argument is the probability  $P_p$ of the statement ``\emph{pfd}$=0$'' being true. So, with probability  $P_p$ the system has zero probability of mishap per demand: we could operate it for an infinite amount of time and a mishap would never happen (remember that we are talking about mishaps due to the design of the self-driving function, not to physical failures, or to fatally reckless behaviour of other drivers, ``acts of God'', etc.: with this restrictive focus, if there is no fault in the system, it will never fail so as to generate a mishap).
With probability $(1-P_p)$, thus, the system does have design faults and will -- sooner or later -- fail: experience a mishap.
The Bayesian description of the problem is that the \emph{pfd} may have any value, with different probabilities:
using the notation we used earlier \cite{strigini_software_2013},
we call $Q$ the unknown \emph{pfd}.
$Q$ is a random variable, with a \emph{prior} probability density function $f_Q(q)$, which in this case is. 

\begin{equation}
\label{eq:genericPosteriorWithPp}
f_{{Q}} \left( q \right) ={P_p}\,\delta \left( q \right) + \left( 1
-{P_p} \right) f_{{{Qn}}} \left( q \right) 
\end{equation}

where $\delta(q)$ is Dirac's delta function and $f_{{{Qn}}} \left( q \right)$ is the probability density function (\emph{pdf}) for the system \emph{pfd} conditional on \emph{pfd}$>0$.

If the future period of operation for which we wish to know that the system is safe enough is made up of $T_{fut}$ demands, the probability of surviving it without mishap (a \emph{reliability} function) is:

\begin{align}
\label{eq:Survival_with_Pp}
R\left(  T_{fut}  \right) 
	=	{\int_{0}^{1}\! \left( 1-q \right) ^{{T_{fut}}}  f_{Q} \left( q \right)  {dq} } \notag \\ 
		=  {\int_{0}^{1}\! \left( 1-q \right) ^{T_{fut}}   (   {P_p}\,\delta \left( q \right) + \left( 1
-{P_p} \right) f_{{{Qn}}} \left( q \right)        )  {dq}  }  \notag \\
= {{P_p} \,+  \, \left( 1-{P_p} \right) \,  \int _{0+}^{1}\!   \left( 1-q \right) ^{{T_{fut}}}   f_{{{Q_N}}} \left( q \right)  {dq}}
\end{align}

One can observe that the reliability will stay higher than $P_p$ for any duration of future operation; the risk we take in the ``gamble'' of operating this system is a weighted sum between zero risk (if  \emph{pfd}$=0$ is true) and the risk due the potential defects in the system.
If we were just concerned with a confidence bound on  \emph{pfd}, we would ignore the fact that with a certain probability those defects are absent and thus pose no risk in operation. That is, we would be needlessly pessimistic.
Once we start operating the system, operation experience feeds new evidence about how much we should trust the hypothesis that \emph{pfd}$=0$. 

\subsection{How no-mishap operation extends the confidence horizon}

After $T_{past}$ independent demands without mishaps, the \emph{posterior} probability of $T_{fut}$ further mishap-free demands is (from Bayes' theorem):
\begin{equation}
\label{eq:genericposterior}
R\left(  T_{fut}  | T_{past}  \right) = 
	\frac
	{\int_{0}^{1}\! \left( 1-q \right) ^{{T_{past}+{T_{fut}}}}  f_{Q} \left( q \right)  {dq} } 
	{\int_{0}^{1}\! \left( 1-q \right) ^{T_{past}}  f_{Q}   \left( q \right)  {dq}  }
\end{equation}

As mentioned earlier, so that the results can be trusted not to be an artefact of unjustified details of the prior distribution, we applied our ``conservative Bayes'' method: given a value of $P_p$, and assuming we do not know the rest of the prior distribution, $f_{{{Q_N}}} \left( q \right)$, our method obtains the most pessimistic posterior reliability, for any pair $\{T_{past}$, $T_{fut} \}$, compatible with that $P_p$ \cite{strigini_software_2013}. Thus our confidence in future mishap-free operation is the most pessimistic given these inputs.
Clearly one could study ``confidence bootstrapping'' with any other choice of prior distribution.
The advantage of our approach is the guaranteed conservatism with respect to uncertainties about the detailed prior distribution of the \emph{pfd}.

\begin{figure}[bhtp!]
	\centering
	\includegraphics[width=0.7\linewidth]{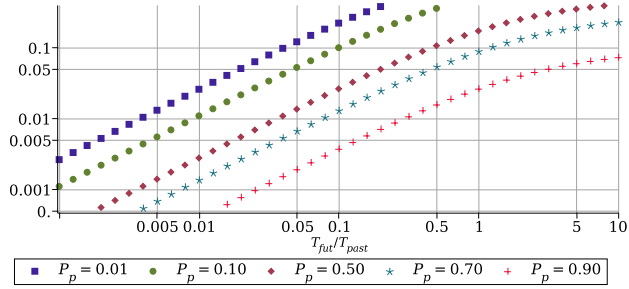} 
	\caption{Probability of one or more mishaps over $T_{fut}$ future demands, supported by given $P_p$ and $T_{past}$ amount of past mishap-free operation.}
	\label{fig:PhorizonSuccess}  
\end{figure}

Figure \ref{fig:PhorizonSuccess} exemplifies the results of the conservative inference.
In this scenario where the prior distribution is described only via its $P_p$ value,  this probability is a function of the ratio $T_{fut}/T_{past}$, so these bi-dimensional plots are sufficient to describe, for any values of $P_p$ and $T_{past}$, the probability of having any mishaps over a certain $T_{fut}$.
Our confidence level is 1 minus that value: the probability of \emph{no} mishap happening in the next $T_{fut}$ amount of operation.
So, given the $P_p$ that one trusts, and the confidence level one desires in future operation without mishaps, one can see how much such future operation can be. We will call the value of $T_{fut}$ for which this confidence holds the ``confidence horizon'' that the experience $T_{past}$ supports for that required confidence level.

E.g., for a required confidence 95\%, the plot shows that if I have $P_p=0.9$ the confidence horizon is about 5 times $T_{past}$ (the exact value of the probability for $T_{fut}=5 \;T_{past}$ is 0.94).
It is useful to visualise the confidence horizon as a multiple of $T{past}$. We will write $T_{hor} = k T_{past}$, so in this case $k=5$; we will use this value in our examples below.

\subsection{Bootstrapping confidence; confidence horizon in amount of operation and in calendar time} 

Our scenario is now as follows. A required probability of no mishaps has been chosen. For this required confidence level, the current $P_p$ value gives a ``confidence extension coefficient'' $k$ that determines the confidence horizon, $T_{hor}=k  T_{past}$.
On this basis, a fleet of vehicles is allowed to operate.
If the fleet size is constant, and the way it is operated (environments in which they run, times of operation, kinds of trips, how many trips of each kind per day or month) does not change, then 
the formulas for the confidence horizon, although written in terms of future number of demands, could just as well be written in terms of time on the road, or calendar time: each one of these measures is proportional to the others. If we use lowercase `t' for calendar time, while keeping uppercase `T' for amounts of operation (e.g., vehicle-days, or total trips) we can write that $t \propto T$. We define an average rate of operation,  ${o_{{avg}}}$ (measured in demands/vehicle/year, i.e., kms/vehicle/year, or trips/vehicle/year, etc, depending on the definition of ``demand'' adopted for measurement), and assume that every vehicle operates at this rate -- an approximation that will become good enough once enough vehicles are on the road (we will discuss its limitations later).
So, at any time, we have clear indication of what our confidence horizon is, and this is growing longer; alternatively, if our main interest is the probability of avoiding mishaps over a fixed horizon, our confidence in this is growing constantly.

But suppose that the vehicle is in production: the fleet size is increasing. A given confidence horizon $T_{hor}=k * T_{past}$ has to be spread over an increasing number of vehicles: in calendar time, my confidence extension coefficient will be less than $k$.
In other words, $t_{hor}/t_{past} < T_{hor}/T_{past}$.
The confidence horizon is a necessary time buffer for decision makers: could the growth of fleet size reduce it too much? Will it progressively shrink to nothing?
We study these questions below.

\subsubsection{Case 1: constant number of vehicles}
At any time ${t}$, with a constant number $n$ of vehicles, $T_{\mathit{past}}={t_{past}} \;  n \;  o_{{avg}}$
and ${t_{hor}} = k   t_{past}$. 

As time passes, the future horizon of confidence expands, proportionally to the time elapsed.
This may be a quite satisfactory situation, e.g. if I am in early pilot operation with a fixed number of vehicles, and want to run this pilot phase for e.g. one year, then, in our example of \textit{k}=5, after 2 months in operation I will have sufficient confidence of mishap-free operation for the remaining 10 months; and this confidence will increase towards 100\%, as is natural, as more time passes without mishaps; or if I am concerned about a fixed population (type Z vehicles, e.g. the set of autonomous bespoke heavy load vehicles in a certain mine), after 1 year I will be confident enough for the next 5 years, and if the planned operating life is - say - 30 years, after 5 years we will be confident enough of safe operation for the rest of the operating life.

With the confidence horizon expanding constantly, if we were interested just in absence of mishap for a constant period into the future, a constant $t_{fut}$, say one year, or 5 years, our confidence in this outcome would keep increasing.
This will remain true for the other scenarios that we examine next.

\subsubsection{Case 2: linear growth of number of vehicles}

Suppose now that after a pilot period of operation, a production line is activated that delivers vehicles at a constant rate  $r_{v}$.
Let us count time from this moment.
The pilot period accumulated a total amount of operation (demands) $T_{p}$,
so that we have ``accumulated confidence'' for a future amount of operation $k T_{p}$,
considered sufficient to decide to start production. As the fleet grows, we can soon ignore this initial "capital" of confidence, as it becomes negligible compared to that accumulated through operation of the mass-produced vehicles.
We assume for the sake of simplicity that every vehicle starts operation as soon as produced, and all operate at the same rate  ${o_{avg}}$.

We want to calculate the confidence horizon after a time $t_{past}$ has elapsed from the start of mass production (and mass operation) of the AV type. The amount of operation until then will be

\newcommand{\ud}{\mathrm{d}}
\begin{equation}\label{eq:quadraticTpast}
T_{past}  =  \int_{0}^{t} (t - \tau)\;{r_{v}}  \,  o_{avg} \; \ud \tau 
=\frac{r_{v} o_{\mathit{{avg}}} t_{past}^{2}}{2}
\end{equation}

giving a confidence horizon 

\begin{equation}\label{(21)}
T_{hor}=k \, T_{\mathit{past}}
\end{equation}

To translate this into calendar time, we consider that 

\begin{equation}\label{(22)}
T_{\mathit{past}}+T_{hor}=\left(k +1\right) T_{\mathit{past}}
\end{equation}

hence, substituting from (\ref{eq:quadraticTpast}):

\begin{equation}\label{(24)}
T_{\mathit{past}}+T_{hor}=\frac{\left(k +1\right) r_{v} o_{\mathit{{avg}}} t_{past}^{2}}{2}
\end{equation}

and observing that, by analogy with  (\ref{eq:quadraticTpast}):

\begin{equation}\label{(25)}
T_{\mathit{past}}+T_{hor}=\frac{r_{v} o_{\mathit{{avg}}} \left( t_{past} +t_{hor}\right)^{2}}{2}
\end{equation}

one sees from equating  (\ref{(24)}) and (\ref{(25)}) that:

\begin{equation}\label{(28)}
T_{hor}=\left(\sqrt{k +1} -1  \right) t_{past} 
\end{equation}

e.g., for our example of \textit{k}=5, $ T_{hor} \approx  1.45 \; t_{past}$.


That is, after 1 year of operation, we have gained the required confidence in future mishap-free operation for approximately another year and a half.

We can call the coefficient on the right hand side 

\begin{equation}\label{(30)}
k_{\mathit{linear}}=  \sqrt{k +1} -1
\end{equation}

We note that:

\begin{itemize}
\item this result does not depend on the production rate, but only on it being constant;
\item though always smaller than $k$, for a high enough $k$ this coefficient $ k_{\mathit{linear}} $, of the order of $\sqrt(k)$,
can still be high, e.g. for 
$ k =10$ , $k_{\mathit{linear}} \approx 3.58$.

\end{itemize}

\normalsize{
 \begin{table}[h!tbp]
        	\centering
        		\begin{tabular}{|c|c|c|}  
        		\toprule
			$P_p$ & $k$ &  $k_{linear}$  \\
		 \midrule
        			0.92  & 5 & 1.45     \\
        			0.82  & 1 & 0.41 \\  
			 0.72 & 0.5  &  0.22 \\ 
			 0.5  & 0.2 & 0.1 \\ 
			 0.1 &  0.04 & 0.02 \\
		\bottomrule
        		\end{tabular}
        	\caption{Some values of $P_p$ and resulting confidence extension coefficients.}
        	\label{tab_notations}
        \end{table}
        }

\subsubsection{Increasing the rate of production or of operation} 
If the production rate increases, the confidence horizon will decrease accordingly (or confidence will decrease if we want the same horizon). E.g., if in the previous example, after five years of operation we  suddenly double the production rate, by opening another identical production line, the amount of past operation becomes, for any $t_{past}=t > 5$, the sum of that due to AVs from the older factory and AVs from the new one:

\begin{equation}\label{(133a)}
T_{{past}}=\frac{r_{v}o_{{avg}}t^{2}}{2}+\frac{r_{v}o_{{avg}}\left(t -5\right)^{2}}{2}
\end{equation}

and
\begin{equation}\label{(133b)}
T_{\mathit{past}}+T_{\mathit{hor}} = 
\frac{r_{v} o_{\mathit{avg}} \left(t +t_{\mathit{hor}}\right)^{2}}{2}+\frac{r_{v} o_{\mathit{avg}} \left(t +t_{\mathit{hor}}-5\right)^{2}}{2}
\end{equation}

but it is also true that 
\begin{equation}\label{(133c)}
T_{\mathit{past}}+T_{\mathit{hor}} = 
\left(k +1\right) T_{\mathit{past}}
\end{equation}

So, substituting 
(\ref{(133a)}) and (\ref{(133b)}) in (\ref{(133c)}) and solving, we obtain, for our example with $k=5$:
\begin{equation}
t_{hor}=-t + 5/2 + \frac{\sqrt{24 \; t^2 - 120 \; t + 275}}{2}
\end{equation}
which at the time the production increases, has the value $5.8$ (units of time): the confidence horizon for the combined output of the two factories is now substantially lower than that calculated without taking into account the new factory, which was $7.2$.

\begin{figure}[t!bhp]
	\centering
	\includegraphics[width=0.7\linewidth]{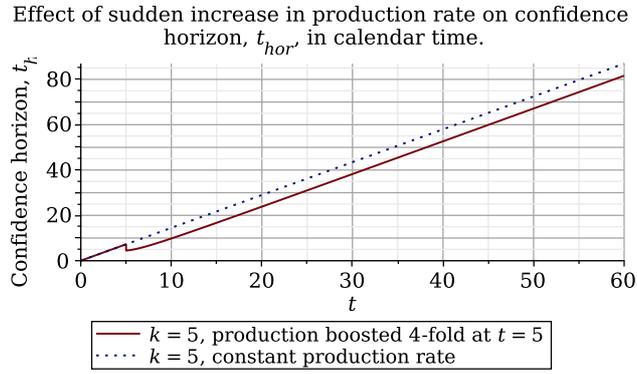}
	\caption{Growth of the confidence horizon in the case of continuous growth of the fleet, and effects of step increase in production rate. After the increase at time $t = 5$, the confidence horizon drops from 11.2 to 4.47, but recovers to the previous value by time $t = 11.0$. }
	\label{fig:figTimeHor}
\end{figure}

\begin{figure}[t!bhp]
	\centering
	\includegraphics[width=0.7\linewidth]{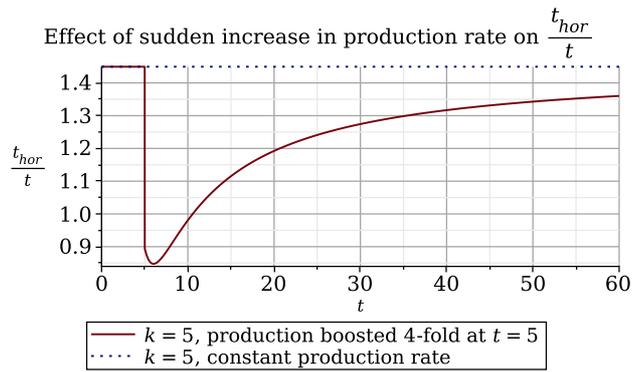}
	\caption{Ratio between confidence horizon and $t_{past}$, constant with continuous growth of the fleet, and the dip following a step increase in production rate.}
	\label{fig:figTimeHorCoef}
\end{figure}

We skip for reasons of space the general solution for this problem, and show an example of these effects in Figs \ref{fig:figTimeHor} and \ref{fig:figTimeHorCoef}, for a greater increment (four-fold) in production rate and thus a larger dip in the confidence horizon. Note that (1) the effects of the dip tend to disappear in the long run; (2) if the increase in production rate is assumed known before it happens, this dip would start, gradually, earlier than shown, as soon as the confidence horizon for those AVs that were produced before the increase reaches the moment at which the increase will take place.

\subsubsection{Maturity and retirement } 
If production continues for a long time, older vehicles will start to be retired, hence the confidence extension coefficient will gradually rise back from $k_{linear}$, to reach $k$ when a constant-fleet size equilibrium is reached (new vehicles are added at the same rate at which older ones are retired).
When production is eventually scaled down and ends, as the fleet size dwindles the confidence horizon will increase even faster, but by then it will generally already extend beyond the end of the life of the fleet.

\section{Discussion}

The discussion until now has assumed that our mathematical model of the real world processes is correct.
We believe that it would be correct in many situations with type-Z AVs (as we defined them in the introduction), and in much more restricted cases for type-A AVs.
We discuss here the various limits to its applicability, what extra evidence or research would be needed to extend these limits, and general insight that can  be derived from this study.

\subsection{The probability of \emph{pfd}$=0$; ``effective fault-freeness''}
\label{subsec_effective_fault-freeness}
The form of conservative Bayesian inference we presented relies on a  prior probability $P_p$ of the system having $0$ \emph{pfd}.
We and our colleagues have developed other forms, but this one has advantages of simplicity; discussing it here will address the essential issues with application to ``confidence bootstrapping''.
\subsubsection{Probability of \emph{pfd}$=0$}
That system \emph{pfd} may be $0$ with non-negligibile probability is plausible for very simple systems, but not generally for the control functions of AVs, which tend to be complex and depend heavily on machine learning.
However, practically all AV manufacturer embrace the principle of including ``safety monitor'' subsystems, which monitor the vehicle's situation checking for any violation of preset safety constraints, and have authority for taking emergency action. These are potentially very simple: for instance, detecting a fixed obstacle via lidar and braking if it is approaching too quickly is a simple function (in principle). For type-Z vehicles, for which the environment is simple and safety  is paramount, it would be plausible that these safety subsystems are simple and so well verified to have a high probability of $0$ \emph{pfd}, making our method applicable.
$P_p$ would never be 1, since subtle misunderstandings of the hazards, or errors in verification, may happen, but it would be high enough to support useful levels of confidence in safety.
We would not claim that this is possible for class-A vehicles, because: (1) the dangerous situations that can develop in traffic may be very complex and not amenable to simple detection or simple accident avoidance decisions; (2) manufacturers are required to pursue a difficult balance between safety and performance (delivering passengers to destination in times that are comparable to human-driven vehicles). However, some manufacturer might try this route of making the safety monitor subsystems more thoroughly verifiable for the whole set of potential hazardous situations.

\subsubsection{``Effective fault-freeness''}
\label{subsec_Effective_fault-freeness}
If the system is complex so that $P_p$ is very small, high quality development and verification could still support a prior confidence about the \emph{pfd} being, if not $0$, at least smaller (better) than a very low bound, $q_L$. We discussed earlier some scenarios in which this could apply \cite{strigini_software_2013}.
The prior confidence has the form
\begin{equation}
\label{eq:spec_of_P_L}
\int_{0}^{q_L}  f_{{Q}} \left( q \right) dq ={P_L}
\end{equation}

leading to a probability of operating over ${T_{fut}}$ without mishap:

\begin{align}
\label{eq:Survival_with_P_L}
R\left(  T_{fut}  \right) =	{\int_{0}^{1}\! \left( 1-q \right) ^{{T_{fut}}}  f_{Q} \left( q \right)  {dq} }  \notag \\
	=	{\int_{0}^{q_L}\! \left( 1-q \right) ^{{T_{fut}}}  f_{Q} \left( q \right)  {dq} }  +  {\int_{q_L}^{1}\! \left( 1-q \right) ^{{T_{fut}}}  f_{Q} \left( q \right)  {dq} } \notag  \\
	\ge  {P_L} \left( 1-q_L \right) ^{T_{fut}}   +  {\int_{q_L}^{1}\! \left( 1-q \right) ^{{T_{fut}}}  f_{Q} \left( q \right)  {dq} } 
\end{align}

If $q_L$ is such that over the time horizon of interest its contribution to risk is negligible (or, equivalently, if we limit our predictions to ${T_{fut}}$ values for which this is true):

\begin{equation}\label{eq:PLusefulness}
(1-q_L)^{T_{fut}} \approx 1
\end{equation}

then (\ref{eq:Survival_with_P_L}) is similar to (\ref{eq:Survival_with_Pp}) in its effect  on prior confidence in mishap-free operation over ${T_{fut}}$, and ability to improve confidence as mishap-free operation accumulates.

For type-A AVs, condition (\ref{eq:PLusefulness}) would probably apply initially only for  very short $T_{fut}$.
But AV manufacturers have ongoing programmes of accompanying road testing and operation of their AVs with continuing verification activities in the form of, e.g., much more extensive simulated driving, pursuing static verification of some safety properties, etc. These activities could progressively improve both  $P_L$ (increasing it) and $q_L$ (reducing it). To what extent this could allow this form of ``confidence bootstrapping'' to remain useful, as the operation amounts involved increase, remains to be studied.

\subsection{Conditions for validity of the model}  \label{subs:BernoulliLimits}
The ``Bernoulli trials'' model we used assumes that the \emph{pfd} will not change.
This implies that neither the vehicle, nor its mode of use, change during operation.
This condition may well be verified for type-Z AVs, meant for well-understood tasks in controlled environments (factories, mines), for which the traditional rule is followed of avoiding change, as far as possible, for critical functions of safety-critical systems.
This rule exists precisely because change undermines the confidence that has been built through expensive verification work.
For type-A AVs, meant for future mass sale to consumers, this condition does not seem to hold at the present state of tumultuous development.
The Bernoulli trial model would only apply over short periods of time between upgrades; although, in the future, maturity of designs and the need to satisfy authorities and consumers about safety might at some point
make changes in self-driving functions much less frequent.

In these conditions of A-type AVs being frequently upgraded during operation, can ``confidence bootstrapping'' work? One way for our model still to help is to convincingly demonstrate that the changes do not reduce (or they actually improve) safety.
Then, using evidence of safe operation of previous versions of a vehicle, as though it concerned the latest version, would only err in the direction of pessimism. 
One of us, with colleagues, has published examples of how to formalise mathematically this kind of argument, and how to take into account the uncertainty that may affect them, for specific scenarios \cite{Littlewood2020_NWTES,Zhao2020_AVfromISSRE_NWTES, ZhaoDSN2021autonomous}. The second paper mentioned also discusses more in depth the conditions for applying Bernoulli trial models.

These methods for accounting for ``changes for the better'' would also apply to changes of the environment of use.
For type-A AVs using public roads, the environment 
will certainly be changing, if nothing else due to the increasing presence of AVs. This may well mean that the environment will gradually become more benign, allowing the method we presented to be extended along the lines of the papers cited above.
Other ways to account for a changing environment are being studied, including monitoring the changes so as to update predictions accordingly \cite{PietrantuonoPopovRusso2020} and making predictions robust by accepting extra conservatism \cite{BishopPovyakalo2017diffProfiles}.

In summary, some extensions to deal with changing environment and evolving AVs are available for specific scenarios, and suggest that research may deliver extensions to a broader range of scenarios.

\subsection{Long-term operation and mishaps} \label{subs:longTerm}

We have described a way of ``bootstrapping confidence'' on the basis of operating an AV without any mishaps.
With this approach, even a single mishap would completely undermine confidence in future operation.
The question of how confidence would grow again with subsequent mishap-free operation has been addressed in other studies, for different contexts \cite{LittlewoodWright1995stoppingRule,ZhaoISSRE2019autonomous}.
 
Regarding our current context, it is reasonable to demand that an acceptably safe AV should not suffer mishaps in early operation: if the target is of the order of less than one serious accident in $10^8$, or more, kms, one such mishap in the first -- say -- 100,000 kms would be a very strong alarm signal.
It is thus reasonable for our method to respond with a total loss of confidence.

So, the method we have described is suitable for a type-Z AV (expected to have no mishaps over its whole lifetime) or the early period of adoption of a type-A AV.

It becomes inadequate later, when a type-A AV, even if acceptably safe, would inevitably start to suffer mishaps.
The fact is that even if an AV is acceptably safe, in that its \emph{pfd} is as required (say, $10^{-r}$), as it goes through extensive operation it will still reach a stage when mishaps are bound to happen.
After the first $10^{r-1}$ demands, there is a non-negligible 10\%  probability of having had at least 1 mishap. After $(0.7 \; 10^r)$ demands, this probability becomes 50\%.
Thus a mishap in the first $10^{r-1}$ demands poses the question whether this AV has acceptable \emph{pfd}$\le10^{-r}$, but hit that 10\% probability of an early  mishap, or instead possibly has an unacceptable \emph{pfd}=$10^{-r+1}$, or even worse \emph{pfd}, but hit a lucky, but not wildly improbable, mishap-free run.
To answer these questions, the method we have presented is as yet inadequate. We plan to study extensions in this directions.
In the much longer run (after a number of demands much larger than $10^r$), the problem disappears: if the rate of mishaps is stable, it is easily assessed with standard statistical methods, to confirm that the vehicle is acceptably safe (or that it is not).

Accidents, or even near misses, that appear due to defects of a system, are likely to trigger attempts to diagnose and remove the defects that cause them, thus -- it is hoped -- improving the \emph{pfd} in subsequent operation.
It has been shown \cite{BishopSafecomp2013ultrareliable} that if these events were due to defects that will be removed, with some probability, following any accidents or near misses, the total number of accidents over the system lifetime will also be bounded. This bound is affected by the number of defects, and by how effective the safety monitors are in causing a defect to be detected before it causes an accident. 
With machine learning systems, just as with mature, very complex conventional systems, it cannot be taken for granted that attempts at removing defects will reduce the \emph{pfd}. 
One cannot exclude that such ``repairs'' will not only be subject to a law of diminishing returns, but possibly just cause the system \emph{pfd} to oscillate up and down without a definite decreasing trend.
However, extending the above model to describe these situations may bring additional insight for this kind of scenarios.

\subsection{Risk criteria}
We have reasoned so far about scenarios in which the main concern is whether there will be any mishap (due to the self-driving functions) in operation. This is one of the possible concerns, appropriate, it would seem, for a public authority that only wants to authorise operation if there is high confidence that the system will not cause harm; or for a manufacturer fearing that any accident during the early life of a new model might turn the public off buying it: accidents would risk all the expected returns on the massive investment made in development. However, there  are other possible viewpoints. For instance, accidents may entail compensation costs after each accident; or recalls after each accident. That is, in some circumstances the dominant concern may not be \emph{whether} there will be accidents, but \emph{how many}.  
This requires an extension to the current model.

Last, while we have focused on how the confidence horizon progressively extends into the future, there will be situations in which the main concern is absence of mishaps over a fixed term into the future. For these circumstances, the model we have presented is very satisfactory, as it shows this confidence increasing steadily as experience of mishap-free operation accumulates.

\section{Conclusions}

``Bootstrapping'' confidence in a system, by operating it on a gradually increasing scale, with the next increase being deemed safe enough on the basis of safe operation in the previous increments, is common practice. With systems that are based on machine learning, hence not amenable to some of the standard ways of gaining confidence in a software-based system before operation, ``bootstrapping'' is even more important.
We have presented a formal mathematical way for a sound (conservative) derivation of \emph{how much} confidence one can really have on the basis of a certain amount of mishap-free operation.

This method applies well in some scenarios, specifically assurance in the short term for the early operation of what we called type-A AVs, those for which assurance is hardest; and probably whole-life assurance for type-Z ones, those built for extremely safe operation in constrained and controlled environments.

For other scenarios, our study is an encouraging indication that similar solutions may be developed, although they may require new research: developing the mathematical methods, but also demonstrating empirically whether the assumptions of these methods hold in practice (within some acceptable degree of approximation), or devising variations in design practices  (e.g. regarding safety monitors), or in data collection practices, that would allow the assumptions to be proved valid and thus grant the benefit of a sounder basis for confidence in future safety.
There are thus various areas for future work. On the mathematical side, the most urgent ones probably concern extending our methods to cover the case in which mishaps do occur (albeit few of them), so filling the gap identified in section \ref{subs:longTerm} between the early, no-mishap days and the very long term phase in which mishaps are rare but numerous enough to make statistical analysis straightforward; and extending the efforts mentioned in \ref{subs:BernoulliLimits} for taking into account changes of the vehicle and/or its environment of use. 

\bibliography{../bibfiles/newLorenzo2021.bib,../bibfiles/DSN2021ref.bib,../bibfiles/perfectionDISPO2.bib,../bibfiles/allrefsLSwithBibtexTweaks.bib,../bibfiles/ISSRE2019_ref.bib}  

\begin{thebibliography}{10}
\providecommand{\url}[1]{#1}
\csname url@samestyle\endcsname
\providecommand{\newblock}{\relax}
\providecommand{\bibinfo}[2]{#2}
\providecommand{\BIBentrySTDinterwordspacing}{\spaceskip=0pt\relax}
\providecommand{\BIBentryALTinterwordstretchfactor}{4}
\providecommand{\BIBentryALTinterwordspacing}{\spaceskip=\fontdimen2\font plus
\BIBentryALTinterwordstretchfactor\fontdimen3\font minus
  \fontdimen4\font\relax}
\providecommand{\BIBforeignlanguage}[2]{{%
\expandafter\ifx\csname l@#1\endcsname\relax
\typeout{** WARNING: IEEEtran.bst: No hyphenation pattern has been}%
\typeout{** loaded for the language `#1'. Using the pattern for}%
\typeout{** the default language instead.}%
\else
\language=\csname l@#1\endcsname
\fi
#2}}
\providecommand{\BIBdecl}{\relax}
\BIBdecl

\bibitem{kalra_driving_2016}
N.~Kalra and S.~Paddock, ``Driving to safety: {How} many miles of driving would
  it take to demonstrate autonomous vehicle reliability?'' \emph{Transp.
  Research Part A: Policy and Practice}, vol.~94, pp. 182--193, 2016.

\bibitem{LittlewoodStriginiCACM93}
B.~Littlewood and L.~Strigini, ``Validation of ultra-high dependability for
  software-based systems,'' \emph{CACM}, vol.~36, no.~11, pp. 69--80, 1993.

\bibitem{ButlerFinelliInfeasibility1993}
R.~Butler and G.~Finelli, ``The infeasibility of quantifying the reliability of
  life-critical real-time software,'' \emph{IEEE TSE}, vol.~19, no.~1, pp.
  3--12, 1993.

\bibitem{zhao_assessing_2020}
X.~Zhao, K.~Salako, L.~Strigini, V.~Robu, and D.~Flynn, ``Assessing
  safety-critical systems from operational testing: {A} study on autonomous
  vehicles,'' \emph{Information and Software Technology}, vol. 128, p. 106393,
  2020.

\bibitem{strigini_software_2013}
L.~Strigini and A.~Povyakalo, ``Software fault-freeness and reliability
  predictions,'' in \emph{Computer {Safety}, {Reliability}, and {Security}},
  ser. {LNCS}, F.~Bitsch, J.~Guiochet, and M.~Kaâniche, Eds., vol. 8153.\hskip
  1em plus 0.5em minus 0.4em\relax Berlin, Heidelberg: Springer Berlin
  Heidelberg, 2013, pp. 106--117.

\bibitem{iec_61508_2010}
\BIBentryALTinterwordspacing
IEC, \emph{{IEC61508}, {Functional} {Safety} of {Electrical}/
  {Electronic}/{Programmable} {Electronic} {Safety} {Related} {Systems}}, 2010.
  [Online]. Available: \url{https://webstore.iec.ch/publication/22273}
\BIBentrySTDinterwordspacing

\bibitem{NUREG6823atwood2003handbook}
C.~Atwood, J.~LaChance, H.~Martz, D.~Anderson, M.~Englehardt, D.~Whitehead, and
  T.~Wheeler, ``Handbook of parameter estimation for probabilistic risk
  assessment,'' U.S. Nuclear Regulatory Commission, Washington, DC, Report
  NUREG/CR-6823, 2003.

\bibitem{strigini_testing_1996}
L.~Strigini, ``On testing process control software for reliability assessment:
  the effects of correlation between successive failures,'' \emph{Software
  Testing, Verification and Reliability}, vol.~6, no.~1, pp. 33--48, 1996.

\bibitem{Zhao2020_AVfromISSRE_NWTES}
\BIBentryALTinterwordspacing
X.~Zhao, K.~Salako, L.~Strigini, V.~Robu, and D.~Flynn, ``Assessing
  safety-critical systems from operational testing: A study on autonomous
  vehicles,'' \emph{Information and Software Technology}, vol. 128, p. 106393,
  2020. [Online]. Available:
  \url{https://www.sciencedirect.com/science/article/pii/S0950584919302356}
\BIBentrySTDinterwordspacing

\bibitem{Littlewood2020_NWTES}
\BIBentryALTinterwordspacing
B.~Littlewood, K.~Salako, L.~Strigini, and X.~Zhao, ``On reliability assessment
  when a software-based system is replaced by a thought-to-be-better one,''
  \emph{Reliability Engineering \& System Safety}, vol. 197, p. 106752, 2020.
  [Online]. Available:
  \url{https://www.sciencedirect.com/science/article/pii/S0951832019301097}
\BIBentrySTDinterwordspacing

\bibitem{ZhaoDSN2021autonomous}
K.~Salako, L.~Strigini, and X.~Zhao, ``Conservative confidence bounds in
  safety, from generalised claims of improvement \& statistical evidence,'' in
  \emph{51st IEEE/IFIP International Conference on Dependable Systems and
  Networks (DSN 2021)}, 2021.

\bibitem{PietrantuonoPopovRusso2020}
\BIBentryALTinterwordspacing
R.~Pietrantuono, P.~Popov, and S.~Russo, ``Reliability assessment of
  service-based software under operational profile uncertainty,''
  \emph{Reliability Engineering \& System Safety}, vol. 204, p. 107193, 2020.
  [Online]. Available:
  \url{https://www.sciencedirect.com/science/article/pii/S0951832020306943}
\BIBentrySTDinterwordspacing

\bibitem{BishopPovyakalo2017diffProfiles}
P.~Bishop and A.~Povyakalo, ``\BIBforeignlanguage{English}{Deriving a
  frequentist conservative confidence bound for probability of failure per
  demand for systems with different operational and test profiles},''
  \emph{\BIBforeignlanguage{English}{Reliability engineering \& system
  safety}}, vol. 158, pp. 246--253, 2017.

\bibitem{LittlewoodWright1995stoppingRule}
B.~Littlewood and D.~Wright, ``Stopping rules for the operational testing of
  safety-critical software,'' in \emph{25th IEEE Annual International Symposium
  on Fault -Tolerant Computing (FTCS-25)}.\hskip 1em plus 0.5em minus
  0.4em\relax Pasadena: IEEE Computer Society Press, 1995, pp. 444--451.

\bibitem{ZhaoISSRE2019autonomous}
X.~Zhao, V.~Robu, D.~Flynn, K.~Salako, and L.~Strigini, ``Assessing the safety
  and reliability of autonomous vehicles from road testing,'' in \emph{2019
  IEEE 30th International Symposium on Software Reliability Engineering
  (ISSRE)}, 2019, pp. 13--23.

\bibitem{BishopSafecomp2013ultrareliable}
P.~Bishop, ``Does software have to be ultra reliable in safety critical
  systems?'' in \emph{Computer Safety, Reliability, and Security}, F.~Bitsch,
  J.~Guiochet, and M.~Ka{\^a}niche, Eds.\hskip 1em plus 0.5em minus 0.4em\relax
  Berlin, Heidelberg: Springer Berlin Heidelberg, 2013, pp. 118--129.

\end{thebibliography}
\bibliographystyle{IEEEtran}
\end{document}